\documentclass{article}
\usepackage{ijcai20}

\usepackage{times}
\usepackage{epsfig}
\usepackage{soul}
\usepackage{url}
\usepackage[hidelinks]{hyperref}
\usepackage[utf8]{inputenc}
\usepackage[small]{caption}
\usepackage{graphicx}
\usepackage{amsmath}
\usepackage{amsthm}
\usepackage{booktabs}
\usepackage{algorithm}
\usepackage{algorithmic}
\usepackage{subfigure}
\usepackage{caption}
\usepackage{multirow}
\usepackage{amsfonts,amssymb}
\title{Accelerating Neural Network Inference by Overflow Aware Quantization}

\author{
	Hongwei Xie\and
	Shuo Zhang\and
	Huanghao Ding\and
	Yafei Song\and
	Baitao Shao\and \\
	Conggang Hu\and 
	Ling Cai\And
	Mingyang Li
\affiliations
	Alibaba Group\\
\emails
\{hongwei.xhw, zs157140, huanghao.dhh, huaizhang.syf, baitao.sbt\\
conggang.hcg, cailing.cl, mingyangli\}@alibaba-inc.com
}

\begin{document}

\maketitle

\begin{abstract}
The inherent heavy computation of deep neural networks prevents their widespread applications. 
A widely used method for accelerating model inference is quantization, by replacing the input operands of a network using fixed-point values. 
Then the majority of computation costs focus on the integer matrix multiplication accumulation.
In fact, high-bit accumulator leads to partially wasted computation and low-bit one typically suffers from numerical overflow. To address this problem,
we propose an overflow aware quantization method by designing trainable adaptive fixed-point representation, to optimize the number of bits for each input tensor while prohibiting numeric overflow during the computation. 
With the proposed method, we are able to fully utilize the computing power to minimize the quantization loss and obtain optimized inference performance.
To verify the effectiveness of our method, we conduct image classification, object detection, and semantic segmentation tasks on ImageNet, Pascal VOC, and COCO datasets, respectively. Experimental results demonstrate that the proposed method can achieve comparable performance with state-of-the-art quantization methods while accelerating the inference process by about 2 times.
\end{abstract}


\section{Introduction}

To date, as a powerful machine learning system architecture, 
deep neural network (DNN) has been applied on numerous applications, e.g., image classification \cite{he2016deep}, object detection \cite{ren2015faster,liu2016ssd}, and semantic segmentation \cite{deeplabv3plus2018}.
However, to achieve high-end performance on complicated problems, most DNN systems require heavy computational resources for model inference, which inevitably limits DNN's deployment on low-cost processors. Such processors are extensively used in billions of commercial products, such as mobile phones, drones, and Internet-of-Things (IoT) devices, which makes DNN acceleration a critical problem in both academia and industry.



\begin{figure}[t]
\subfigure[For a 32-bit accumulator, one instruction can compute 4 multi-adds with 128-bit register.]{
\begin{minipage}{8cm}
\centering
\includegraphics[scale=0.36]{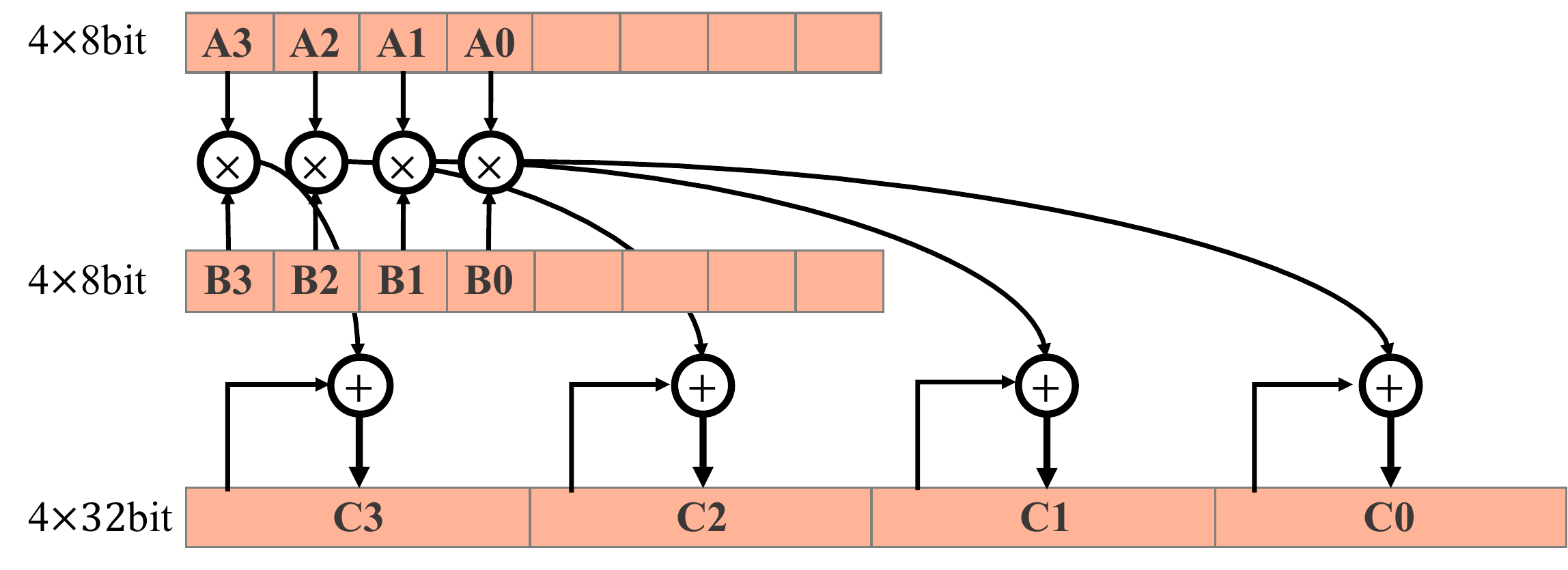}
\label{fig:vmlal_s16}
\end{minipage}
}
\subfigure[For a 16-bit accumulator, one instruction can compute 8 multi-adds at the same time.]{
\begin{minipage}{8cm}
\centering
\includegraphics[scale=0.36]{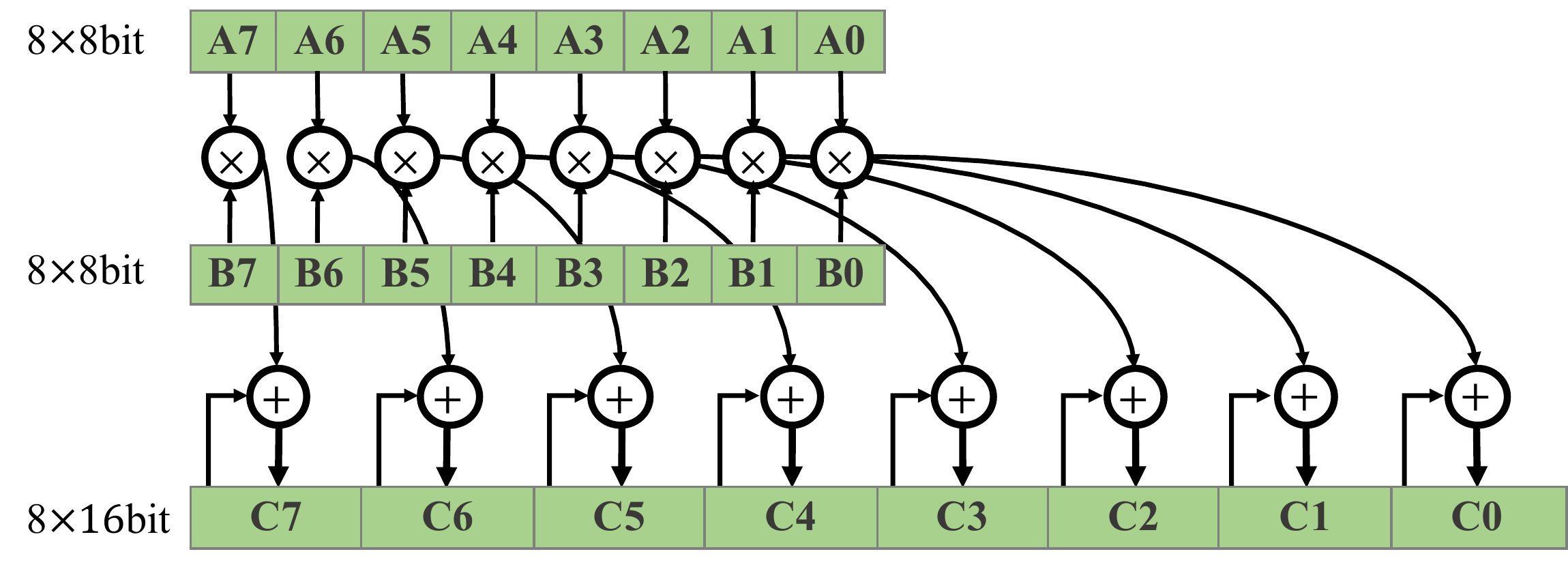}
\label{fig:vmlal_s8}
\end{minipage}
}
\caption{A representative example to show that replacing a 32-bit accumulator with a 16-bit one leads to a double amount of multiply-adds operations at the same time.
\label{fig:teaser}
}
\end{figure}

To allow DNN acceleration, researchers have developed various approaches, which can be roughly divided into two groups. 
The first group of methods focus on designing or searching for more compact and efficient network structures, which allow for reduced number of parameters and computations while achieving comparable performance. 
Representative methods in this category include MobileNet \cite{howard2017mobilenets}, EfficientNet \cite{tan2019efficientnet}, ProxylessNAS \cite{cai2018proxylessnas}, and so on.
The second group aims at improving the efficiency of arithmetic computation, i.e., the multiply-accumulate (MAC) operation, as it dominates most computations during the DNN model inference.
One widely used method is to approximate the original floating-point calculation using fixed-point operation to achieve computation acceleration. This type of method is well-known as \textit{quantization} \cite{jacob2018quantization}.
Representative visualization of quantization is shown
in \autoref{fig:vmlal_s16}, where 8-bit fixed-point integers are used to approximate floating-point values and 32-bit fixed-point variables are used to hold MAC results. Moreover,
in addition to speeding up the MAC operations, quantization also achieves better parallel computing based on the capability of modern CPUs. 


By comparing \autoref{fig:vmlal_s8} against \autoref{fig:vmlal_s16}, it can be shown that if 16-bit fixed-point variables are used to hold the MAC result, the degree of parallelism will be doubled and the I/O times will be halved.
However, when 16-bit holder is used, numerical overflow on MAC results becomes a frequently-happening problem that must be explicitly considered.
A straightforward solution is to use low-bit quantization ($<$8-bit) for all operands, which however leads to loss of quantization precision and significantly reduced the performance.
In addition, low-bit quantization still needs to take up more physical bits (e.g., 4-bit quantization still requires 8-bit physical operands) in most modern CPUs, making the computational resources partially wasted. 

To summarize, existing quantization methods utilize fixed number of bits to represent float values, while both high-bit and low-bit representation have limitations. The former suffers from numerical overflow problems and the latter one leads to model precision degradation.
To tackle this problem, we propose a novel method to adaptively determine the quantization precision in DNNs, by optimizing the number of bits for operands while prohibiting overflow on the low-bit MAC result holders. 
To achieve this, we introduce 
a trainable quantization range mapping factor $\alpha$ into each layer of a DNN network, which automatically scales the quantized result to prevent the undesirable overflow.
In addition, we propose a quantization-overflow aware training framework for learning the quantization parameters, to minimize the performance loss caused by post quantization \cite{krishnamoorthi2018quantizing}.
To verify the effectiveness of our method, we conducted tests on a couple of state-of-the-art light-weighted DNNs for a variety of tasks on different benchmarking datasets. Specifically, our experiments include image classification, object detection, and semantic segmentation, which are tested on ImageNet, Pascal VOC, and COCO datasets, respectively. 
Experimental results demonstrate that, compared with state-of-the-art quantization methods, the proposed method can achieve comparable performance while speeding up the inference efficiency by about 2 times.

The main contributions of this paper are listed as follows:
\begin{enumerate}
  \item We propose an overflow aware quantization (OAQ) algorithm for accelerating DNNs, that is able to adaptively maximize the number of bits for operands while prohibiting the numeric overflow. 
  \item To ensure optimized performance, we design a quantization overflow aware training framework (QOAT), to automatically learn the parameters used by the proposed OAQ algorithm. 
  \item We conduct extensive experiments on three public datasets using state-of-the-art light-weight DNNs.  The results verify the effectiveness of our method on a variety of tasks including image classification, object detection, and semantic segmentation.
\end{enumerate}

\section{Related Work}
In this work, we focus on quantization methods that accelerate inference on off-the-shelf  hardware platforms.
While non-uniform quantization methods \cite{stock2019and,Gao2019BeyondPQ} are also shown to be effective, they do not allow efficient implementing on modern CPUs.

A representative early method is binary quantization \cite{rastegari2016xnor}, which quantizes both weights and activations to one bit, by using bit-shift and bit-count instead of multiply-adds operators to speed up. This method achieves acceptable performance on common over-parameterized networks, like AlexNet \cite{krizhevsky2012imagenet}, but leads to substantial performance degradation on light-weight networks, e.g. ResNet-18 \cite{he2016deep} and MobileNet \cite{howard2017mobilenets}. 
As of today, one of the most widely used quantization methods is 8-bit quantization  \cite{jacob2018quantization,krishnamoorthi2018quantizing,tqt2019}, which is extensively applied in different applications on various hardware platforms. 8-bit quantization converts the inference process into integer-only operations, that could result in \(2\times\sim3\times\) faster inference process on mobile CPUs. 
However, when tacking resource-demanding large networks or deployed on resource-constrained platforms, additional DNN acceleration is still required.

To allow further DNN acceleration, low-bit quantization techniques are under active exploration, in which a key problem is to 
balance the inference speed and model performance.
\cite{choi2018pact} proposes PACT to train not only the weights but also the clipping parameters for clipped ReLU using gradient descent.
\cite{louizos2018relaxed} presents RQ to optimize the quantization part with gradient descent.
However, both methods suffer from performance degradation on lightweight networks. In fact,
when quantizing both weights and activations to 4 bits, PACT \cite{choi2018pact} leads to accuracy reduction from 70.9\% to 62.44\%. Also,
quantizing MobileNet using RQ \cite{louizos2018relaxed} to 6-bit achieves 68\% accuracy only. Additionally, existing  low-bit techniques only focus on the classification task.
Evaluation results on other popular tasks, e.g., object detection or semantic segmentation, are limited on literatures.
Inference accelerating using properties of processors is another research direction.
\cite{zeng2019kcnn} proposes to decrease computational load of multiplications, by exploiting the parallel computing capability of modern CPUs.
\cite{gong2019differentiable} implements 2-bit fast integer arithmetic with ARM NEON technology and achieves \(1.7\times\) speed up over NCNN \cite{ncnn}. However, the reported results still suffer from significant performance loss, e.g., 4-bit quantized MobileNet-v2 leads to performance drop from 71.8\% to 64.8\%.

Moreover, there are a number of mixed-precision quantization methods \cite{wang2019haq,wu2018mixed,dong2019hawq}, which focus on searching for an optimal bit-width setup that can achieve high-level acceleration on customized hardware platforms while avoiding performance drop. 
Compared to those methods, the proposed OAQ framework focuses on off-the-shelf devices, and addresses the numerical overflow problem by designing trainable parameters in each layer of a network. 


\section{Quantization Prerequisites}
In this section, we first present the general algorithmic framework of quantization methods. Subsequently, we analyze the inherent overflow problem of the general framework and provide mathematical conditions that allow overflow aware quantization (OAQ). Detailed steps on performing OAQ approach are discussed in the next section. 

\subsection{Standard Quantization Method}


To convert a floating-point real number $r \in \mathbb{R}$ to a fixed-point quantized number \(q \in \mathbb{Z}\), the following affine mapping function is typically used \cite{jacob2018quantization}:
\begin{equation}
    r=S(q-Z), \label{eq:quant_affine}
\end{equation}
where $S$ is the scale factor and $Z$ is the zero-point parameter.
By denoting $q_a^{(i,k)}$ the element at $i$th row and $k$th column of matrix $A$, the matrix multiplication of $C=A\times B$ in quantized domain can be computed element-wise as:
\begin{equation}
    q_c^{(i,k)}=Z_c+P\sum_{j=1}^{N}\left((q_a^{(i,j)}-Z_a)(q_b^{(j,k)}-Z_b)\right). \label{eq:basic_mat_mul}
\end{equation}
where the multiplier $P$ is defined as
\begin{equation}
    P=\dfrac{S_aS_b}{S_c},
\end{equation}
which can be implemented by fixed-point multiplication and efficient bit-shift \cite{jacob2018quantization}.
By expanding terms in~(\ref{eq:basic_mat_mul}), we are able to obtain:
\begin{equation}
\begin{aligned}
    q_c^{(i,k)}=Z_c+P( NZ_aZ_b-Z_aM_b^{(k)} \\ 
    -Z_bM_a^{(i)}+\sum_{j=1}^{N}q_a^{(i,j)}q_b^{(j,k)}), \label{eq:opt_mat_mul}
\end{aligned}
\end{equation}
where
\begin{equation}
    M_b^{(k)}=\sum_{j=1}^{N}q_b^{(j,k)}, M_a^{(i)}=\sum_{j=1}^{N}q_a^{(i,j)}. \label{eq:quant_acc}
\end{equation}
In~(\ref{eq:opt_mat_mul}), the majority of computation costs are the core integer matrix multiplication accumulation:
\begin{equation}
    \sum_{j=1}^{N}q_a^{(i,j)}q_b^{(j,k)}. \label{eq:macs}
\end{equation}
To compute for (\ref{eq:macs}), the standard method is to accumulate products of 8-bit values (signed or unsigned) with a 32-bit integer accumulator (also see \autoref{fig:vmlal_s16}):
\begin{equation}
    C_{\in \mathbb{Z}_{32}}\quad +=\quad A_{\in \mathbb{Z}_{8}}\quad \times \quad B_{\in \mathbb{Z}_{8}}. \label{eq:i32_i8_i8}
\end{equation}
where $\mathbb{Z}_{y}$ represents the space of $y$-bit representable number.
As a result, a 32-bit register is required for caching the intermediate result.
A NEON instruction can compute multiple multiply-adds at the same time, but it is limited by the register size and the number of multiplying and summing units on board. The limited register resource is usually the main bottleneck. 
Moreover,
we note that under most ARM architectures, there is no instruction to implement (\ref{eq:i32_i8_i8}). 
To this end, existing popular mobile inference engines (e.g., TFLite and NCNN) typically rely on 
\begin{equation}
    C_{\in \mathbb{Z}_{32}}\quad +=\quad A_{\in \mathbb{Z}_{16}}\quad \times \quad B_{\in \mathbb{Z}_{16}}. \label{eq:i32_i16_i16}
\end{equation}
as a replacement, i.e., VMAL.S16 on ARM architecture.

Additionally, we note that loading data between register and memory is heavy operations. By applying (\ref{eq:i32_i16_i16}), more register space is used to implement a bigger micro-kernel$\footnote{\url{https://engineering.fb.com/ml-applications/qnnpack/}}$. This largely reduces the frequency of transferring intermediate results between register and memory. 
To show more details, we evaluated the performance of convolutions on MTK8167s CPU with different implementations.
Using VMLAL.S8 and 4x8 micro-kernel, the computation efficiency is improved by 36\%. 
While applying a 4x16 micro kernel, we achieved 73\% speed up.

\subsection{Optimize Quantization Operations}
To optimize the efficiency of current quantization scheme, we seek to use 16-bit integer as accumulator
\begin{equation}
C_{\in \mathbb{Z}_{16}}\quad +=\quad A_{\in \mathbb{Z}_{8}}\quad \times \quad B_{\in \mathbb{Z}_{8}}. \label{eq:i16_i8_i8}
\end{equation}
Compared to (\ref{eq:i32_i8_i8}), one NEON instruction here (VMLAL.S8 on ARM architecture) is able to compute double amount of multiply-adds operations, as shown in \autoref{fig:vmlal_s16} and \autoref{fig:vmlal_s8}.

However, by directly using (\ref{eq:i16_i8_i8}), numerical overflow becomes an unavoidable problem. This is one of the core problem we seek to resolve in this work. To make this possible, we first re-expand (\ref{eq:basic_mat_mul}) by 
assuming \(q_a\) as the quantized inputs and \(q_b\) as the quantized weights:
\begin{equation}
\begin{aligned}
    q_c^{(i,k)}=Z_c+P\sum_{j=1}^{N}\left((q_a^{(i,j)})(q_b^{(j,k)}-Z_b) \right.\\
    \left. -Z_aq_b^{(j,k)}+Z_aZ_b\right), \label{eq:basic_mat_mul_1}
 \end{aligned}
\end{equation}
which can be rewritten as 
\begin{equation}
    q_c^{(i,k)}=Z_c+P\left[\sum_{j=1}^{N}\left(q_a^{(i,j)}\hat{q}_b^{(j,k)}\right)+B\right], \label{eq:basic_mat_mul_3}
\end{equation}
where
\begin{equation}
    B=-\sum_{j=1}^{N}Z_aq_b^{(j,k)}+NZ_aZ_b,
\end{equation}
\begin{equation}
    \hat{q}_b^{(j,k)}=(q_b^{(j,k)}-Z_b).
\end{equation}
In above equations, \(Z_a, Z_b, q_b\) are all be constant values once training is complete, and thus \(B\) and \(\hat{q}_b^{(j,k)}\) can be computed in advance to improve inference efficiency.

Based on above equations, we point out that, to allow efficient computation under (\ref{eq:i32_i8_i8}), the following three conditions must be satisfied:
\begin{equation}  
\left\{  
    \begin{array}{l}
        q_a^{(i,j)} \in \mathbb{Z}_{8}  \\  
        \hat{q}_b^{(j,k)} \in \mathbb{Z}_{8}  \\  
        \sum_{j=1}^{N}q_a^{(i,j)}\hat{q}_b^{(j,k)} \in \mathbb{Z}_{16}   
    \end{array}  ,
\right.
\label{eq:conditions}
\end{equation}  
It is important to note that the last condition in (\ref{eq:conditions}) should hold for both final number and all intermediate numbers. We also point out that, the second and last conditions in (\ref{eq:conditions}) are {\em not} always naturally true. Without taking special consideration, numerical overflow will frequently happen. To guarantee (\ref{eq:conditions}) in a DNN system, additional algorithms need to be designed and implemented.

\section{Overflow Aware Quantization Framework}


This section describes the details of our overflow aware quantization algorithm, including both representation and training, to ensure the important three conditions in  (\ref{eq:conditions}).
\subsection{Adaptive Integer Representation}




One straightforward method to reduce accumulation overflow is to narrow the range of each quantized value.
For example, by using 4-bit quantization instead of 8-bit quantization, real values are mapped to \([-8,7]\) instead of \([-128,127]\).  
As a result, numerical overflow becomes significantly less likely to happen, at a cost of wasting a large number of bits and reducing accuracy.
To this end, we propose an adaptive float-bit-width method to fully utilize the  representation capability without arithmetic overflow.

\begin{figure}[t]
\begin{center}
\includegraphics[width=0.85\linewidth]{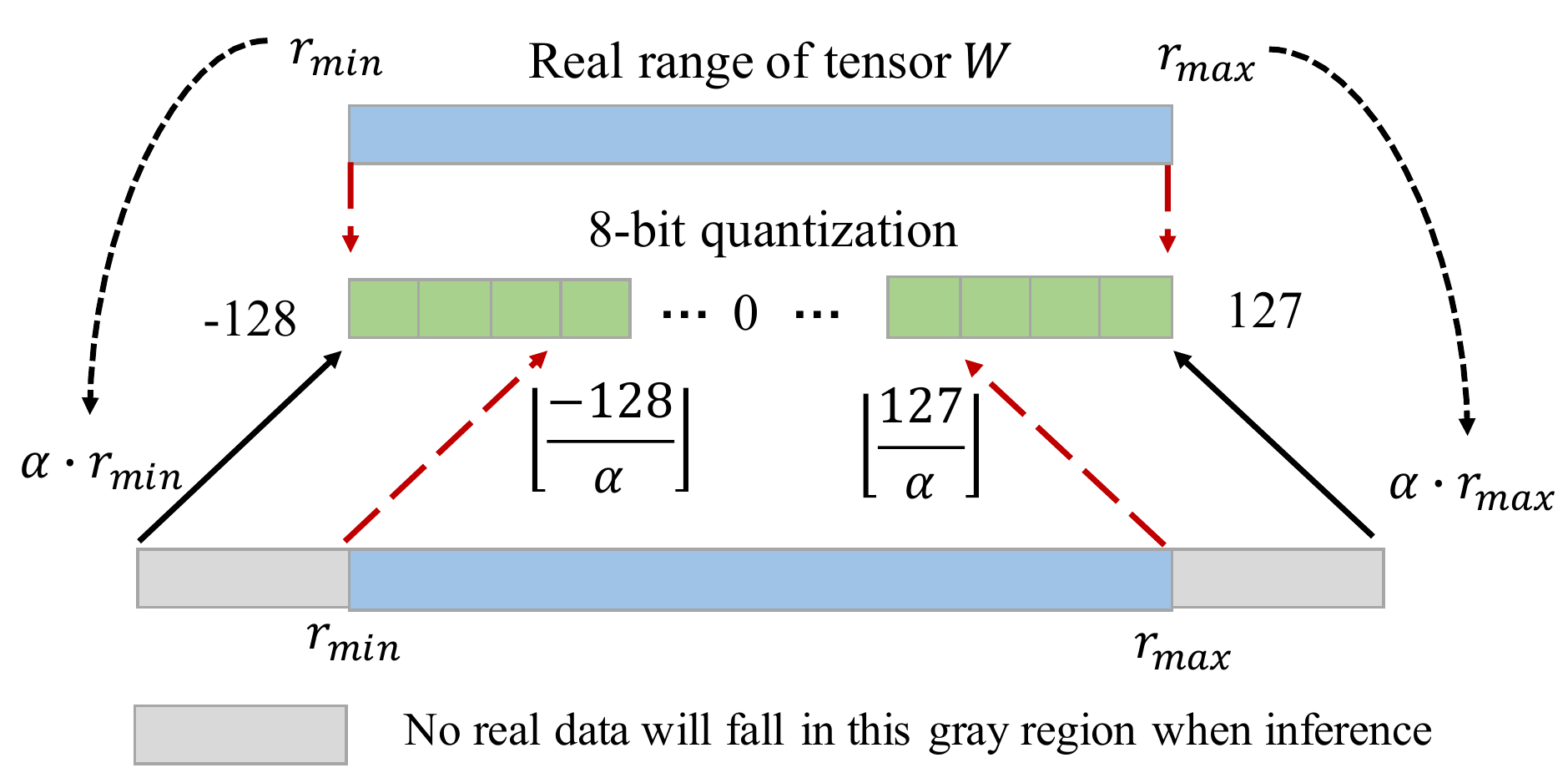}
\end{center}
   \caption{We introduce a float factor $\alpha$ to adjust the affine relationship between the real range and quantized range, e.g., enlarging $\alpha$ to narrow down the quantized value range.}
\label{fig:scale_minmax}
\end{figure}

Specifically, we use a float quantization range mapping factor \(\alpha\) to adjust the affine relationship between the real range and quantized range, as shown in \autoref{fig:scale_minmax}.
Scaled by \(\alpha\), the original 8-bit (the biggest bit-width can be used) quantization range \([-128,127]\) is mapped to \(\left[\lfloor \frac{-128}{\alpha} \rfloor, \lfloor \frac{127}{\alpha} \rfloor \right]\). 
By enlarging \(\alpha\), we are able to narrow down the quantized value range until the arithmetic overflows are eliminated.

To present this mathematically, the affine function (\ref{eq:quant_affine}) can be written as
\begin{equation}
    q=\frac{r}{S}+Z.
\end{equation}
By applying the scale factor $\alpha$ on \(S\)
\begin{equation}
\begin{aligned}
{S'} & = \alpha \cdot S \\
     &= \alpha \cdot \frac{r_{max}-r_{min}}{2^b-1} ,
\end{aligned}
\end{equation}
the quantized value is narrowed to
\begin{equation}
    q'=\frac{r}{S'}+Z,
\end{equation}
where \(r_{min}\) and \(r_{max}\) are the minimum and maximum limits of the real value, and $b$ is the number of bits for the quantized value.

We name this as float-bit-width method since it differs from the traditional integer low-bit representation whose quantization ranges have to be chosen from the limited bit-width set, e.g, 3-bit for \([-4,3]\) or 4-bit for \([-8,7]\). 
By using the proposed method, it is feasible to utilize different
quantization range mapping factor \(\alpha\) in {\em each} layer, to maximize the representation capability while prohibiting overflow. Additionally, since \(\alpha\) is continuous, it can also be easily integrated into training process of DNNs. Details on
adaptively learning \(\alpha\) for weights and activations of each layer will be discussed in the next subsection.

In addition to the range of quantized value, the value distribution is also critical.
We expect the quantized values to be centered and gathered around zeros. As a result, The accumulated number will be more likely to be away from overflow. 
Initializing weights with a normal-distributed-like initializer and applying L1-L2-Normalization are representative methods that can be used.
\begin{figure}[t]
\begin{center}
\includegraphics[width=0.85\linewidth]{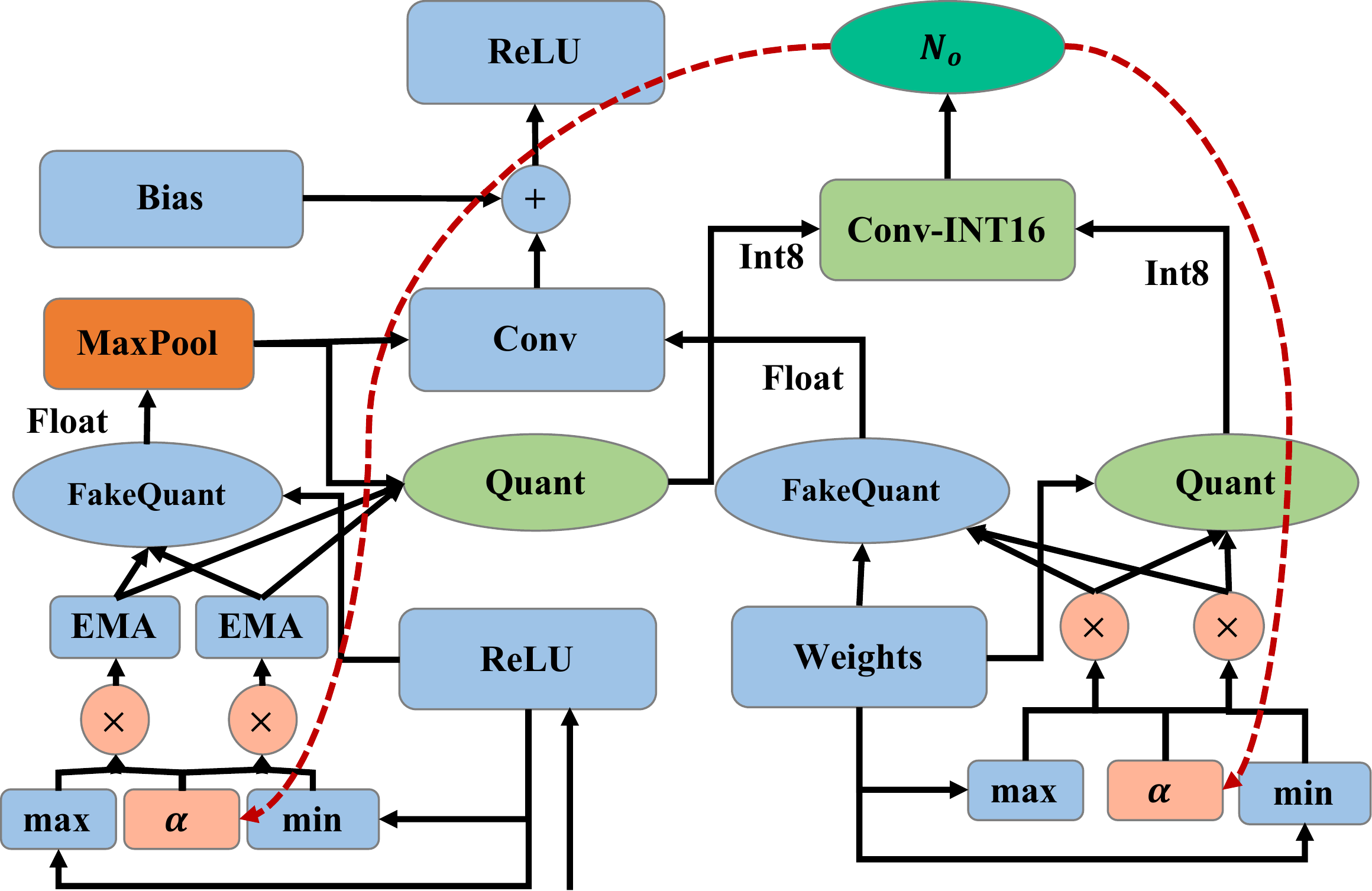}
\end{center}
  \caption{We insert Quant nodes into each layer of computation graph to calculates the amount of arithmetic overflow.}
\label{fig:overflow-aware-framework}
\end{figure}
\subsection{Learning Quantization Range Mapping Factor}


To find proper \(\alpha\) for each layer to simultaneously prohibit arithmetic overflow and retain the model's performance, we propose a quantization-overflow aware training framework.
As shown in \autoref{fig:overflow-aware-framework}, inspired by simulating quantization effects in forwarding pass \cite{jacob2018quantization}, we add the overflow aware module that simulates neural operations, e.g., Convolution, FullyConnect using 16-bit accumulator for capturing arithmetic overflow in the forward pass.

In the forward pass, 8-bit quantization simulates quantized inference by implementing in floating-point arithmetic which is called FakeQuantization (\(Q_{fake}\)) \cite{jacob2018quantization}. 
To adaptively learn \(\alpha\), capturing the amount of arithmetic overflow in the quantized inference process is required. 
Therefore, we insert a Quantization node (\(Q_{real}\)) into each layer of the computation graph in addition to \(Q_{fake}\). 
\(Q_{real}\) requires \(r_{min}\) and \(r_{max}\) to produce 8-bit quantized values which are the same with inference engine.
\(r_{min}\) and \(r_{max}\) are scaled by \(\alpha\) before being passed into \(Q_{real}\) and \(Q_{fake}\).
\(r_{min}\) and \(r_{max}\) in activations are aggregated by exponential moving averages (EMA) with the smoothed parameter close to 1, such that the observed ranges are smoothed, allowing the network to enter a more stable state.
The created convolution operation with 16-bit accumulator Conv-INT16 takes the 8-bit quantized values as input to simulate the integer-only-inference on the inference engine. This calculates the amount of arithmetic overflow  \(N_{o}\) that accumulates in the inference process, including all types of overflow defined in the overflow-free conditions (\ref{eq:conditions}).
An easy way to capture overflow signals from the popular training framework, e.g., TensorFlow or PyTorch is to compare the results between the regular 32-bit convolution and the 16-bit convolution. Certainly, you can make a more efficient implementation by some low-level language like C++.
As the process of getting \(N_{o}\) is integer-only computation, gradient descent becomes not a proper method in back-propagation.
To address this problem, we use a simple rule to compensate for this.

Specifically, when \(N_{o}\) is bigger than zero, increasing \(\alpha\) by:
\begin{equation}
    \alpha\mathrel{+}=min(lr_{i}*{\rm log}(N_{o}),l_c),
\end{equation}
where 
$l_c$ is the fixed maximum learning rate. Additionally, \(lr_{i}\) is the dynamic learning rate for increasing \(\alpha\), which decays with the steps of training. After a large number of iterations, \(lr_i\) is decayed to a quite small value to stabilize the training state.

Alternatively, if \(N_{o}\) is zero, we decrease \(\alpha\) by:
\begin{equation}
    \alpha\mathrel{-}=lr_{d},
\end{equation}
where \(lr_{d}\) is the learning rate for decreasing \(\alpha\), whose properties and physical representation are both similar to \(lr_{i}\).
For improving training efficiency, calculating \(N_{o}\) and updating \(\alpha\) are executed every \(M\) steps, e.g., 10 or 50 during iterations.

The strategy of inserting \(Q_{real}\) is slightly different from \(Q_{fake}\). As shown in \autoref{fig:overflow-aware-framework}, fake quantization for input or output is inserted after the activation function or a bypass connection, e.g. adds or concatenates as they change the real min-max ranges. 
For the operations such as max-pooling, up-sampling or padding, \(Q_{fake}\) is not required.
But the 16-bit neural operation, e.g., Conv-INT16 only accepts quantized integer inputs. The outputs of \(Q_{real}\) has to be directly passed into it. That means \(Q_{real}\) must be inserted ahead of Conv-INT16 in the computation graph.
Finally, \(Q_{real}\) and \(Q_{fake}\) share the same \(r_{min}\), \(r_{max}\) and \(\alpha\), for strictly simulating the same inference process.

\section{Overflow Study}


\begin{figure}[t]
	\centering
	\subfigure[The activation $\alpha$ of each layer in MobileNet-v1 trained on ImageNet. 
	]{
		\begin{minipage}{0.85\linewidth}
			\centering
			\includegraphics[scale=0.4]{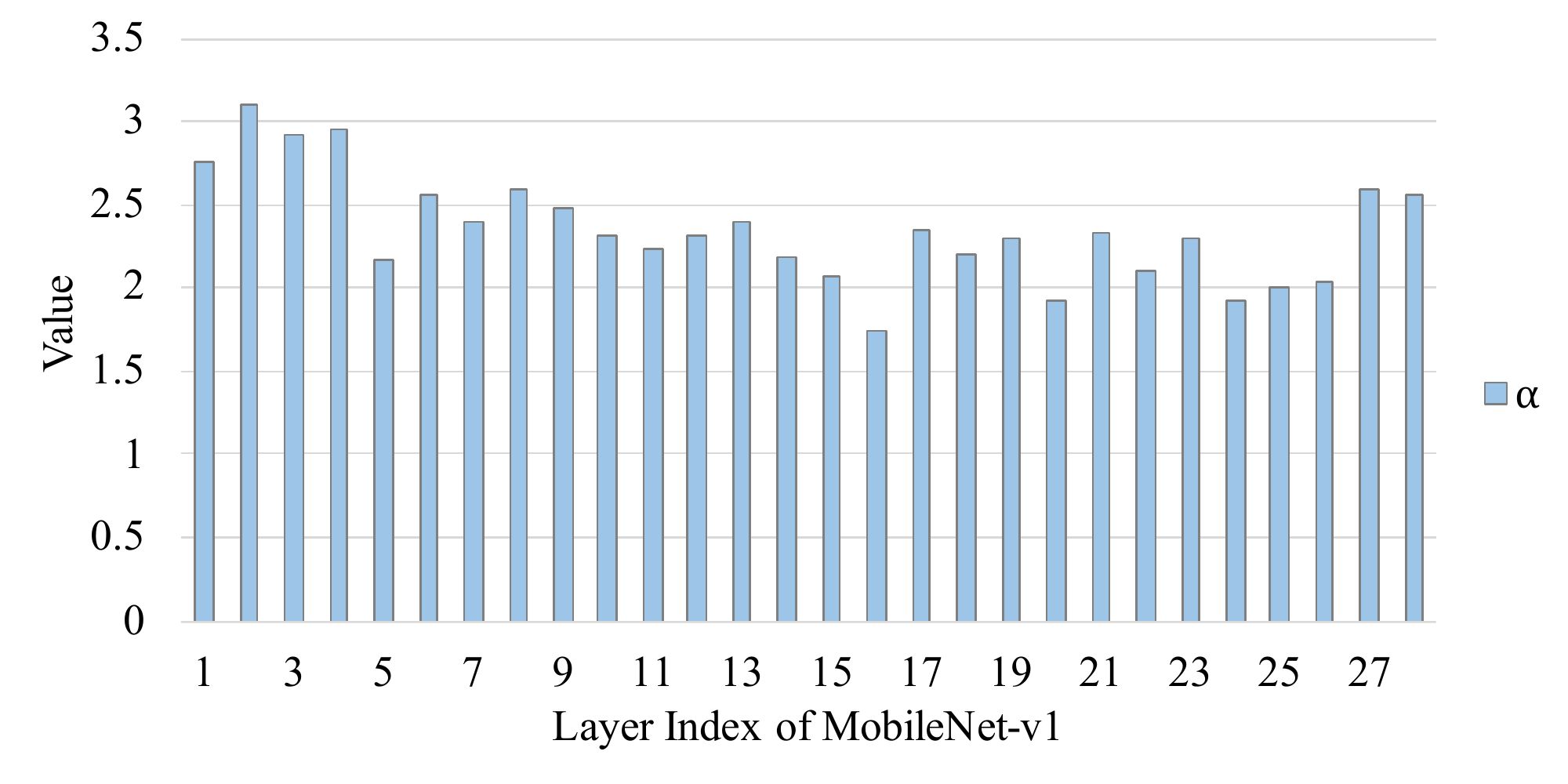}
			\label{fig:alpha1}
		\end{minipage}
	}
	\subfigure[The activation $\alpha$ of each layer in MobileNet-v1-SSD trained on Pascal VOC. 
	]{
		\begin{minipage}{0.85\linewidth}
			\centering
			\includegraphics[scale=0.4]{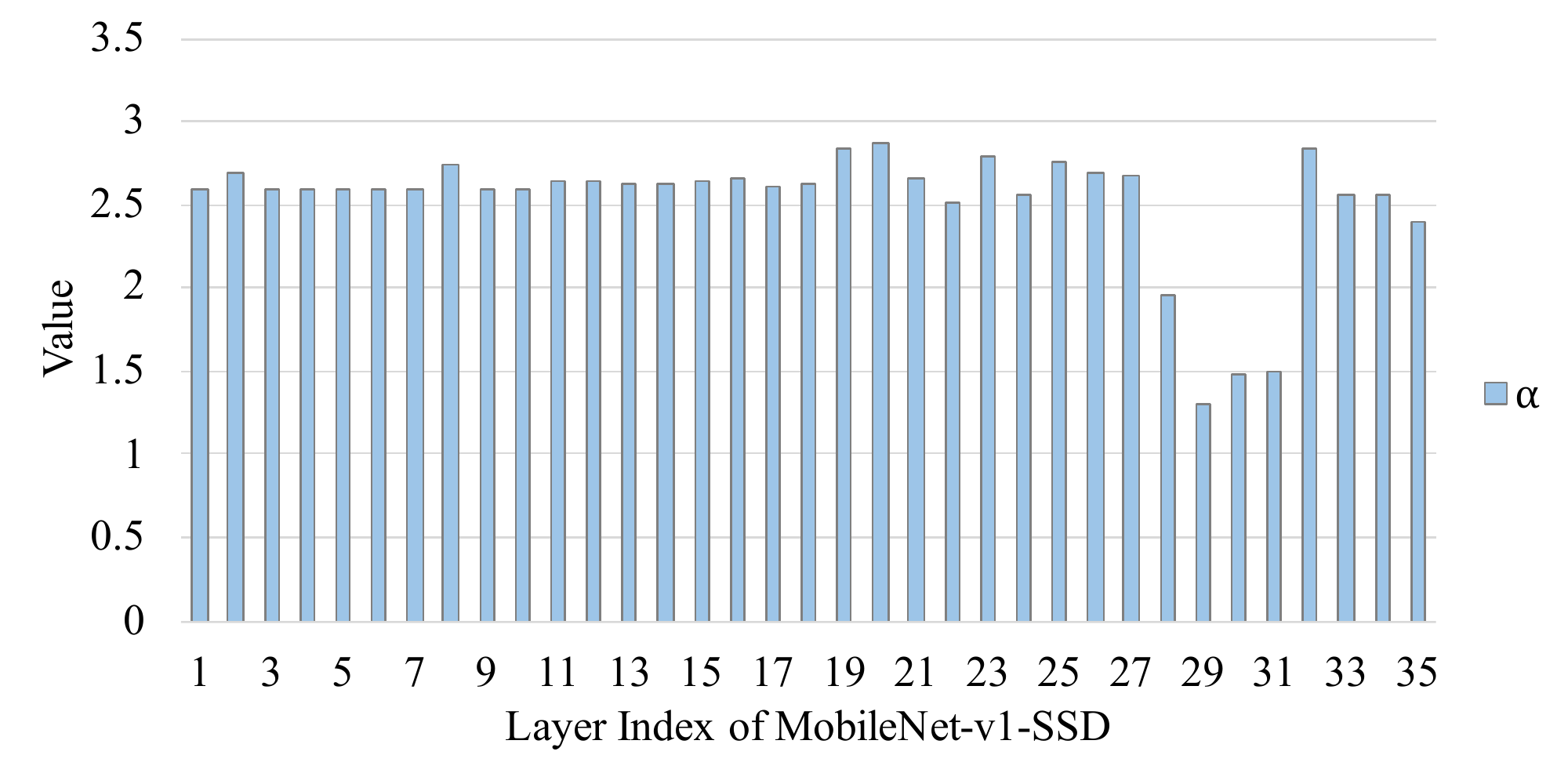}
			\label{fig:alpha2}
		\end{minipage}
	}
	\caption{Provides two representative results of per-layer factor values, implying that the quantized values are mostly between 6bits and 7bits.
		\label{fig:per_layer_bits}
	}
\end{figure}
In this section, we first show in-depth analysis on the proposed framework of adaptive scale parameter and explain why we focus on comparing OAQ against state-of-the-art 6-bit quantization methods. 
Subsequently, we study the overflow sensitive of different neural network models and show how overflow ratio affecting the overall performance.

\subsection{Per-Layer Adaptive Scale}
Since our key algorithmic contribution is a method to adaptively learn the quantization range mapping factor $\alpha$ for DNN's each layer, it is also important to demonstrate the distribution of $\alpha$ across networks in our experiments.
\autoref{fig:per_layer_bits} provides two representative results of per-layer $\alpha$ values, adaptively trained by using the proposed  quantization-overflow aware training framework. 
From this figure, we can observe that the computed scale factor varies across layer, emphasizing our `scale-per-layer' adaptive design. Additionally, we note that the majority of activation factor $\alpha$ fall into the range of $[2,4]$, indicating that the quantized values are mostly between 6bits and 7bits.

\begin{figure}[t]
	\begin{center}
		\includegraphics[width=0.95\linewidth]{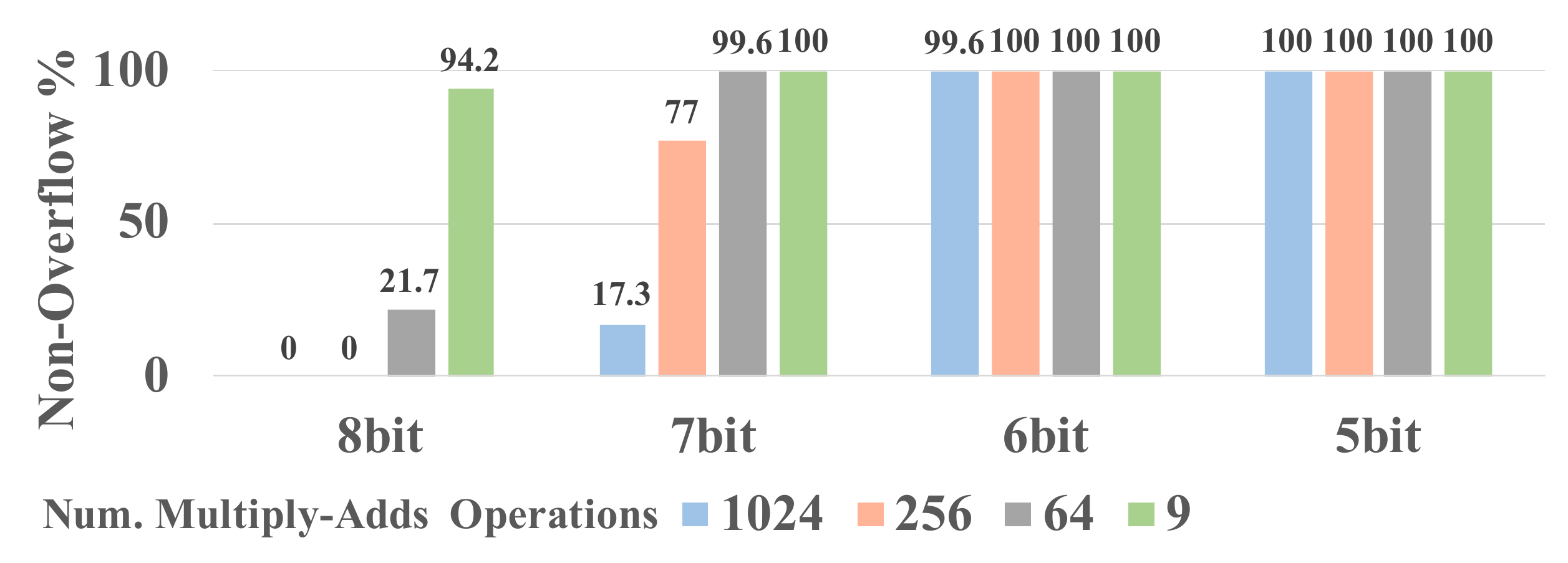}
	\end{center}
	\caption{The Non-Overflow ratio of randomly simulated 9, 64, 256, 1024 operands' multiply-adds 100,000 times with different quantization bits and a 16-bit accumulator.}
	\label{fig:overflow_rate}
\end{figure}

Additionally, we estimate the overflow ratio of low-bit quantization with a 16-bit accumulator by random multiply-adds simulation. 
Specifically, we uniformly sampled $N$ values from the quantization value range and tried to capture an overflow signal. The signal is computed by applying consecutive multiply-adds operators of the $N$ numbers on a 16-bit holder. Since 3x3 depth-wise convolution and 1x1 point-wise convolution are widely used in light-weight DNN architectures, we chose $N$ to be \{9, 64, 256, 1024\} representatively. Subsequently, the rough Non-Overflow ratio was calculated by 100000 independent simulation runs. 
As shown in \autoref{fig:overflow_rate}, 6-bit quantization could retain overflow free in most cases, while 7-bit and 8-bit methods are risky.

Both \autoref{fig:per_layer_bits} and \autoref{fig:overflow_rate} indicate that at most 6-bits could be used if we apply low bit quantization to address the overflow problem. Therefore, we will focus on comparing OAQ against state-of-the-art 6-bit quantization methods.

\subsection{Overflow Sensitive Study}
\begin{figure}[t]
\subfigure[Inject overflow into some layers(L0: the first layer, L1: the second layer, L26: the penultimate layer, L28: The last layer before Softmax, ALL: inject overflow into all layers) of MobileNet-v1 and evaluate on ImageNet.]{
\begin{minipage}{8cm}
\centering
\includegraphics[scale=0.35]{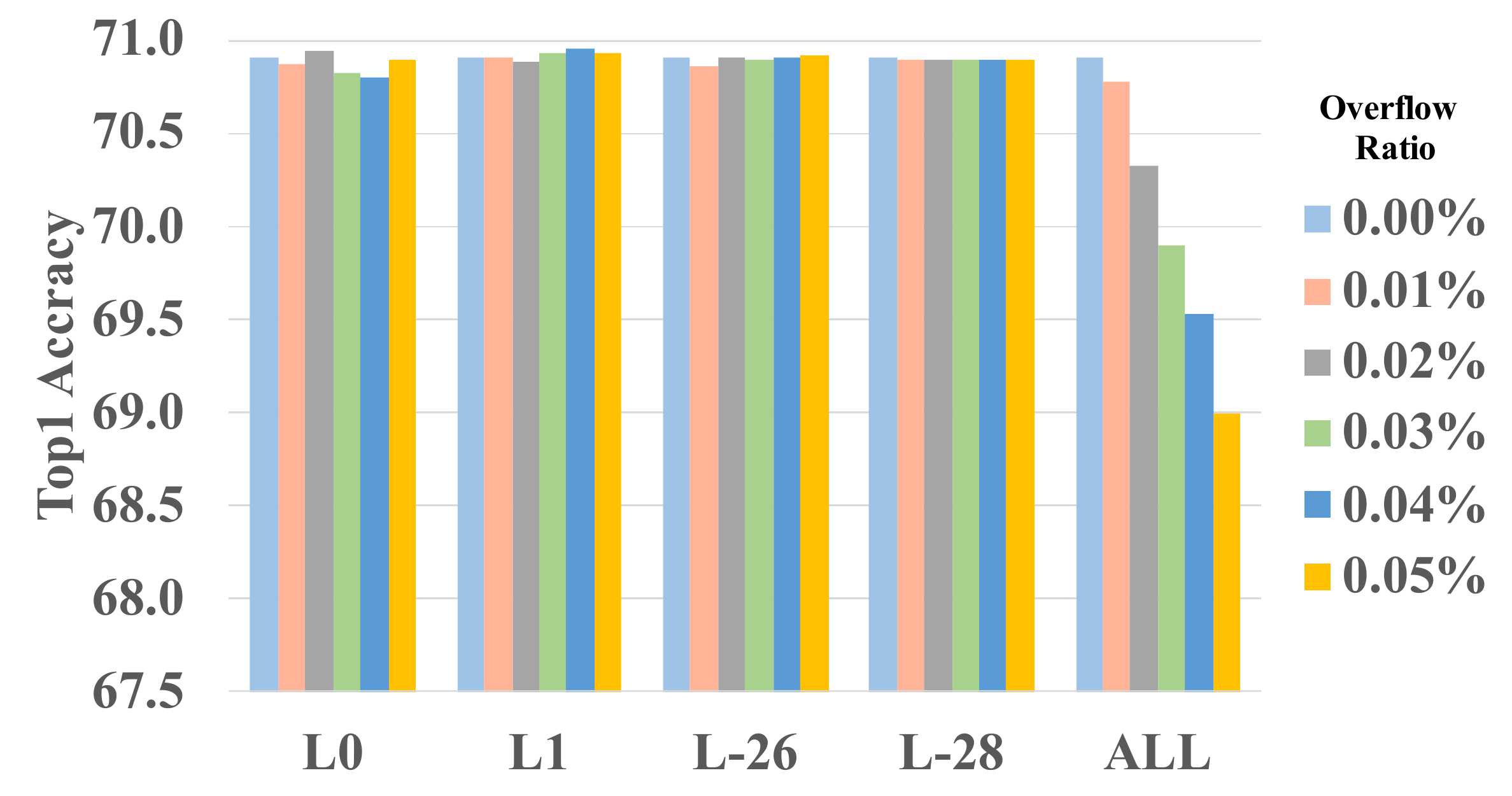}
\label{fig:overflow_mv1}
\end{minipage}
}
\subfigure[Inject overflow into some layers(L0: the first layer, L1: the second layer, LFs: the 6 feature layers that provide inputs to the last headers , LLs: The last 12 headers that regressing bounding boxes and classification scores) of MobileNet-v2-SSD and evaluate on COCO.]{
\begin{minipage}{8cm}
\centering
\includegraphics[scale=0.35]{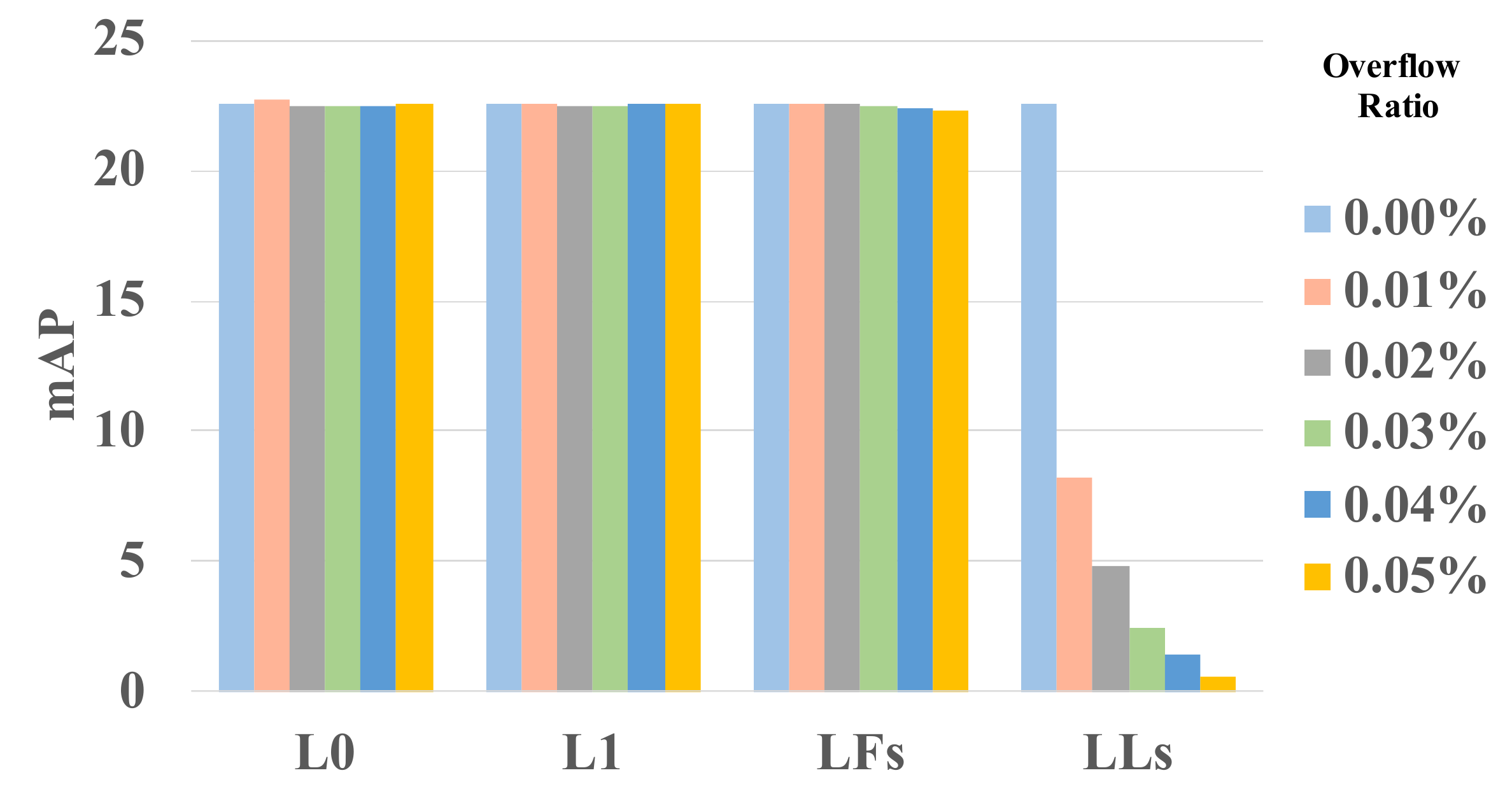}
\label{fig:overflow_mv2_ssd}
\end{minipage}
}
\caption{Inject arithmetic overflow into different layers of neural network, e.g., MobileNet-v1 and MobileNet-v2-SSD. In classification model, even 0.05\% arithmetic overflow won't damage the overall accuracy. But in detection model, 0.01\% overflow will result in the model unavailable.
\label{fig:overflow_study}
}
\end{figure}

Someone still worry about that the quantization range mapping factors learned from training data may be not applied to all test data safely. For example, a little arithmetic overflow may cause the entire model to fail. Actually, we indeed found some overflow (usually on the order of tens after quantization-overflow-aware training) while inference on test data, but the outputs seem still right.
In order to analyze the impact of overflow quantitatively, we designed the overflow simulation experiments. Specifically, we manually inject overflow to the output of some layers to see their influence on the final result. And we tested the overflow sensitive of MoblieNet-v1 on ImangeNet and MobileNet-v2-SSD on COCO Detection Challenge respectively. The results are shown in \autoref{fig:overflow_study}.
In classification model, inject 0.05\% arithmetic overflow into any one layer has no impact on the overall accuracy. When apply it to all layers, the accuracy slowly drops along with the growth of overflow ratio. In detection model, overflow on the last layers  affect the final result seriously. 0.01\% overflow will cause the model unavailable. But the detection model can keep high-performance when injecting overflow into other layers.

\section{Experiments}
To demonstrate the performance of our proposed OAQ framework, we evaluate both inference accuracy/recall and run-time characteristics on representative public benchmarking datasets.

\subsection{ImageNet}
The first experiment is to benchmark MobileNet-v2 \cite{sandler2018mobilenetv2}, MoibleNet-v1 \cite{howard2017mobilenets}, and MobileNet-v1 with depth-multiplier 0.25 on the ILSVRC-2012 ImageNet. This dataset consists of 1000 classes, 1.28 million training images, and 50K validation images. 
We fine-tuned MobileNet models from pre-trained model zoo (TF-slim$\footnote{\url{https://github.com/tensorflow/models/tree/master/research/slim}}$).
All of those models were re-trained for 20 epochs.
During training, the inputs were randomly cropped and resized to 224x224 before being fed into the network.
Since the inputs of the first layer are not suitable for scaling, we skipped learning the quantization range mapping factor of the first layer's weights. This rule was applied to all the following experiments, including PACT \cite{choi2018pact}, which always uses 8-bit weights in the first layers.
We report our 
 evaluation results using Top-1 and Top-5 accuracy.

In our tests, we focus on comparing OAQ against state-of-the-art 6-bit quantization methods, including PACT \cite{choi2018pact}, RQ \cite{louizos2018relaxed}, and SR+DR \cite{gysel2018ristretto}. Comparison against other selective state-of-the-art methods were also conducted.
As shown in \autoref{tb:imgnet}, OAQ even outperforms 8-bit Quantization-Aware Training (QAT) \cite{jacob2018quantization} in certain cases.
From Table \ref{tb:imgnet}, we observe that when quantizing MobileNet-v1, OAQ outperforms RQ \cite{louizos2018relaxed} and SR+DR\cite{gysel2018ristretto} by large margins.
Although PACT \cite{choi2018pact} is better on Top-5, the proposed method consistently performs better in other metrics, especially on MobileNet-v1-0.25, i.e., 1.36\% higher than PACT.
We also take DSQ \cite{gong2019differentiable} into comparison, as it also achieved \(1.7\times\) speed up over NCNN on an ARM Cortex-A53 CPU. The result shows, on MobileNet-v2 our results are significantly better, e.g., 6.84\% higher. 
\begin{table}[t]
  \center
  \begin{tabular}{l@{\hspace{2mm}}l@{\hspace{2mm}}l@{\hspace{2mm}}c@{\hspace{2mm}}c}
  \toprule
  Model & Method & Bits & MAC bits & Top1 / Top5 \\
  \midrule
  \multirow{5}*{\shortstack{MobileNet\\-v2}}
  & FP  &  32 & 32 & 71.80 / 91.00 \\
  & QAT & ~~8 & 32 & 70.90 / 90.00 \\
  \cline{2-5}
  &PACT & ~~6 & 32 & \underline{71.25} / \underline{90.00} \\
  &DSQ  & ~~4 & 32 &64.90 / ~~—~~~ \\
  &Our  & adaptive & 16 & \textbf{71.64} / \textbf{90.10}\\
  \midrule
  \multirow{6}*{\shortstack{MobileNet\\-v1}}
  & FP    &  32 & 32 & 70.90 / 89.90 \\
  & QAT   & ~~8 & 32 & 70.10 / 88.90 \\
  \cline{2-5}
  & PACT  & ~~6 & 32 & \underline{70.46} / \textbf{89.59} \\
  & RQ    & ~~6 & 32 & 68.02 / 88.00\\
  & SR+DR & ~~6 & 32 & 66.66 / 87.17 \\
  & Our   & adaptive & 16 & \textbf{70.87} / \underline{89.56}\\
  \midrule
  \multirow{4}*{\shortstack{MobileNet\\-v1-0.25}}
  & FP   &  32  & 32 & 49.80 / 74.20 \\
  & QAT  & ~~8  & 32 & 48.00 / 72.80 \\
  \cline{2-5}
  & PACT & ~~6  & 32 & \underline{46.03} / \underline{70.07} \\
  & Our  & adaptive & 16 & \textbf{47.38} / \textbf{72.14} \\
  \bottomrule
  \end{tabular}
  \caption{Evaluating on ImageNet and comparing against SOTA low-bit quantization methods.}
  \label{tb:imgnet}
\end{table}

\subsection{Detection on VOC and COCO}

\begin{table}[t]
 \begin{tabular}{l@{\hspace{2mm}}l@{\hspace{2mm}}l@{\hspace{2mm}}c@{\hspace{2mm}}c}
 \toprule
 Model & Method & Bits & MAC bits & mAP \\
 \midrule
 \multirow{4}*{\shortstack{MobileNet-v1\\SSD}}&FP&32&32&73.83 \\
 &QAT&8&32&72.54 \\
 \cline{2-5}
 &PACT&6&32&70.88 \\
 &Our&adaptive&16&\textbf{72.53}\\
 \midrule
 \multirow{4}*{\shortstack{MobileNet-v2\\SSDLite}}&FP&32&32&72.79 \\
 &QAT&8&32&72.02 \\
 \cline{2-5}
 &PACT&6&32&70.50 \\
 &Our&adaptive&16&\textbf{71.84}\\
 \bottomrule
 \end{tabular}
 \caption{Evaluating on Pascal VOC Detection Challenge and comparing with 6bit-PACT.}
 \label{tb:voc_detection}
\end{table}

 \begin{table}[t]
 \begin{tabular}{l@{\hspace{2mm}}l@{\hspace{2mm}}l@{\hspace{2mm}}c@{\hspace{2mm}}c}
 \toprule
 Model & Method & Bits & MAC bits & mAP \\
 \midrule
 \multirow{4}*{\shortstack{MobileNet-v1\\SSD}}&FP&32&32&23.7 \\
 &QAT&8&32&23.0 \\
 \cline{2-5}
 &PACT&6&32&18.4 \\
 &Our&adaptive&16&\textbf{22.0}\\
 \midrule
 \multirow{4}*{\shortstack{MobileNet-v2\\SSDLite}}&FP&32&32&22.7 \\
 &QAT&8&32&21.4 \\
 \cline{2-5}
 &PACT&6&32&18.1 \\
 &Our&adaptive&16&\textbf{21.8}\\
 \bottomrule
 \end{tabular}
 \caption{Evaluating on COCO Detection Challenge and comparing with 6bit-PACT.}
 \label{tb:coco_detection}
\end{table}

To illustrate the applicability of our method to object detection, we applied OAQ on MobileNet-v1-SSD \cite{howard2017mobilenets,liu2016ssd} and MobileNet-v2-SSDLite \cite{sandler2018mobilenetv2} and evaluated on the 2012 Pascal VOC object detection challenge and 2017 MSCOCO detection challenge. We implemented our experiments with TensorFlow and fine-tuned models from TensorFlow Object Detection API$\footnote{\url{https://github.com/tensorflow/models/blob/master/research/object_detection/g3doc/detection_model_zoo.md}}$ (only backbone since the SSD header differs with tasks). 
We first trained them with 32-bit floating-point precision to achieve state-of-the-art performance, and subsequently fine-tuned on VOC for \(40,000\) steps with batch size 32 and \(60,000\) steps on COCO with batch size 48 respectively.

The results on VOC and COCO are listed in \autoref{tb:voc_detection} and \autoref{tb:coco_detection} respectively. On both of VOC detection challenge and COCO detection challenge, the proposed method outperforms the 6-bit PACT \cite{choi2018pact} significantly. In addition, our method achieves comparable performance with 8-bit QAT \cite{jacob2018quantization} and is of small mAP drop from the original model, i.e., about 1\% mAP drop in VOC and less than 2\% mAP drop in COCO.

\subsection{Segmentation on VOC}
To demonstrate the generalization of our method to semantic segmentation, we applied OAQ on DeepLab \cite{deeplabv3plus2018} with MobileNet-v2 backbone (depth-multiplier 0.5 and 1.0). The performance was evaluated on the Pascal VOC segmentation challenge, which contains 1464 training images and 1449 validation images.
The results are shown in \autoref{tb:voc_segment}. When quantizing the original model to 6 bits with PACT \cite{choi2018pact}, there is a significant drop in performance, e.g., MobileNet-v2-dm0.5 backbone dropped 10.9\% in mIOU. By comparison, the proposed OAQ method achieved comparable performance with QAT, and only drop 1.8\% and 0.4\% on MobileNet-v2-dm0.5 and MobileNet-v2 backbone respectively compared to the original model.
\begin{table}[t]
 
 \center
  \begin{tabular}{l@{\hspace{2mm}}l@{\hspace{2mm}}l@{\hspace{2mm}}c@{\hspace{2mm}}c}
  \toprule
  Model & Method & Bits & MAC bits & mIOU \\
  \midrule
  \multirow{4}*{\shortstack{MobileNet-v2\\dm0.5\\deeplab}}&FP&32&32&71.8 \\
  &QAT&8&32&70.4 \\
  \cline{2-5}
  &PACT&6&32&60.9 \\
  &Our&adaptive&16&\textbf{70.0}\\
  \midrule
  \multirow{4}*{\shortstack{MobileNet-v2\\deeplab}}&FP&32&32&75.3 \\
  &QAT&8&32&74.8 \\
  \cline{2-5}
  &PACT&6&32&70.4 \\
  &Our&adaptive&16&\textbf{74.9}\\
  \bottomrule
  \end{tabular}
  \caption{Evaluating on Pascal VOC Segmentation Challenge and comparing with 6-bit PACT.}
 \label{tb:voc_segment}
 \end{table}
 
\begin{table}[t]
  \begin{tabular}{l@{\hspace{1mm}}l@{\hspace{1mm}}c@{\hspace{1mm}}c}
  \toprule
  CPU & Inference Engine & \shortstack{MobileNet-v1} &\shortstack{ResNet18} \\
  \midrule
  \multirow{4}*{\shortstack{Allwinner \\V328}}
  & TFLite & 550 & 1370 \\
  & MNN    & 469 & 1605 \\
  & MNN + OAQ & \textbf{341} & \textbf{1021}\\
  & Ours + OAQ & \textbf{277} & \textbf{895}\\
  \midrule
  \multirow{5}*{\shortstack{MTK8167s}}&TFLite&387&950 \\
  & NCNN       & 351 & 706 \\
  & MNN        & 311 & 916 \\
  & MNN + OAQ  & \textbf{220} & \textbf{604}\\
  & Ours + OAQ &\textbf{189} & \textbf{585}\\
  \bottomrule
 \end{tabular}
 \caption{Comparison on inference time (msec) using MobileNet-v1 and ResNet18 networks.}
  \label{tb:ben_v1}
 \end{table}

\subsection{Inference Efficiency Benchmark}

Finally, we demonstrate the capability of DNN acceleration of the proposed method on different low-cost hardware platforms.
Specifically, 
we benchmarked computational efficiency 
on two selectively platforms, i.e., Allwinner V328 and MTK8167, 
whose processor architectures are ARM-Cortex-A7 and ARM-Cortex-A35 respectively. 
In this test, MobileNet-v1 and ResNet-18 were used as representative DNN models to conduct inference. To run the DNN inference,
three popular neural network inference engines for low-cost platforms were selected, i.e., TFLite, MNN \cite{alibaba2020mnn}, and NCNN, under single-threaded implementation within one core. 

The experimental results are listed in Table \ref{tb:ben_v1}, which clearly demonstrate that the proposed method outperforms competing methods by wide margins. In fact, on both
MTK8167s and Allwinner V328, our method achieves \(2\times\) faster runtime than TFLite and \(1.85\times\) faster than NCNN.



\section{Conclusion}
In this paper, we propose an overflow aware quantization method to allow significant DNN inference time acceleration, and minimize the loss of accuracy. To achieve this, we propose to adaptively adjust the number of bits used for representing quantized fixed-point integers. 
This scheme is also incorporated into a novel training framework, to adaptively learn the overflow-free quantization range while maintaining high-end performance. 
By using the proposed method, an extremely light-weight neural network can achieve comparable performance with the 8-bit quantization method on the ImageNet classification challenge.
Comprehensive experiments were also conducted to verify that our method can also be applied to various dense prediction tasks, e.g., object detection, and semantic segmentation by outperforming competing state-of-the-art methods.



\bibliographystyle{named}
\bibliography{egbib}

\end{document}